\documentclass[lettersize,journal]{IEEEtran}
\usepackage{amsmath,amsfonts}
\usepackage{algorithmic}
\usepackage{algorithm}
\usepackage{array}
\usepackage[caption=false,font=normalsize,labelfont=sf,textfont=sf]{subfig}

\usepackage{hyperref}
\hypersetup{
 colorlinks=true, 
 linkcolor=black, 
 filecolor=black, 
 urlcolor=black,  
 citecolor=red, 
}
\usepackage{booktabs}

\usepackage{graphicx}
\usepackage{color}
\usepackage[cmyk]{xcolor}

\usepackage[numbers]{natbib}

\usepackage{amsmath}
\usepackage{multirow}
\usepackage{amssymb}
\usepackage{textcomp}
\usepackage{stfloats}
\usepackage{url}
\usepackage{verbatim}
\usepackage{graphicx}
\usepackage{cite}
\begin{document}

\title{Dynamic Link Prediction for New Nodes in Temporal Graph Networks}

\author{
Xiaobo Zhu, Yan Wu*, Qinhu Zhang*, Zhanheng Chen, Ying He
\thanks{
This work was supported by the National Natural Science Foundation of China under Grant NO. 62002266 and 62002297.

Xiaobo Zhu, Yan Wu and Ying He are with the College of Electronic and Information Engineering, Tongji University, Caoan Road 4800, Shanghai 201804, China. (e-mail: xiaobozhu@tongji.edu.cn, yanwu@tongji.edu.cn, heyingzy@163.com )

Qinhu Zhang is with EIT Institute for Advanced Study, Tongxin Road No.568, Ningbo, Zhejiang, China, 315201 (e-mail: qinhuzhang@eias.ac.cn)

Zhanheng Chen is with Department of Clinical Anesthesiology, Faculty of Anesthesiology, Naval Medical University, Shanghai 200433, China. (e-mail: chen\_zhanheng@163.com)



%

}}

\markboth{Journal of \LaTeX\ Class Files,~Vol.~14, No.~8, Jun~2021}%
{Shell \MakeLowercase{\textit{et al.}}: A Sample Article Using IEEEtran.cls for IEEE Journals}

\IEEEpubid{0000--0000/00\$00.00~\copyright~2021 IEEE}

\maketitle
\begin{abstract}
Modelling temporal networks for dynamic link prediction of new nodes has many real-world applications, such as providing relevant item recommendations to new customers in recommender systems and suggesting appropriate posts to new users on social platforms. Unlike old nodes, new nodes have few historical links, which poses a challenge for the dynamic link prediction task. Most existing dynamic models treat all nodes equally and are not specialized for new nodes, resulting in suboptimal performances. In this paper, we consider dynamic link prediction of new nodes as a few-shot problem and propose a novel model based on the meta-learning principle to effectively mitigate this problem.  Specifically, we develop a temporal encoder with a node-level span memory to obtain a new node embedding, and then we use a predictor to determine whether the new node generates a link. To overcome the few-shot challenge, we incorporate the encoder-predictor into the meta-learning paradigm, which can learn two types of implicit information during the formation of the temporal network through span adaptation and node adaptation.
The acquired implicit information can serve as model initialisation and facilitate rapid adaptation to new nodes through a fine-tuning process on just a few links. Experiments on three publicly available datasets demonstrate the superior performance of our model compared to existing state-of-the-art methods.

\end{abstract}


\section{Introduction}
\IEEEPARstart{I}{n} recent years, temporal graph networks have attracted considerable attention from researchers\citep{r49, r50} due to their ability to reveal intricate evolutionary patterns at the microscopic level \citep{r44}. A salient feature of temporal networks is that newly arriving nodes typically have few initial links to other nodes\citep{r3}. Considering the widespread existence of such new nodes or cold-start nodes in real-world scenarios including providing relevant item recommendations to new customers in recommender systems and suggesting appropriate posts to new users on social platforms. it is worthwhile to study the prediction of links for nodes with only a few observed links. However, there is very little in existing temporal models specialized to deal with this. 

There are two challenges in dynamic link prediction of new nodes. The first is how to design an effective temporal model to capture the embedded representation of new nodes. 
\IEEEpubidadjcol 
Recent research for node embeddings in temporal network has made technical advances, moving from discrete-time dynamic graphs \citep{r4} to continuous-time dynamic graphs \citep{r8,r9}. These methods have moved from only being able to represent the coarse-grained evolutionary patterns of dynamic graphs \citep{r6,r7} to accurately representing network evolution at a fine level, resulting in impressive performance in node prediction and classification tasks \citep{r10,r11}. Despite these advances, current dynamic GNNs treat all nodes equally and do not specifically target new nodes \citep{r41}. The prediction loss associated with new nodes may be overshadowed by the overall loss, making the training parameters shared by all nodes in the dynamic GNN model unsuitable for sparsely connected nodes \citep{r43}. In addition, new nodes have fewer historical interactions than general nodes, inhibiting the model's ability to capture essential preference information about the nodes. As a result, these models exhibit suboptimal performance in predicting the future links of new nodes. 

 The second challenge is how to make the temporal model learn more effective information with a small number of links? Meta-learning \citep{r13,r14,r5} has gained widespread attention in recent years as a prominent technique for solving few-shot problems. To address the problem of few-shot graphs, researchers have combined state-of-the-art meta-learning approaches with GNN techniques to achieve superior performance. The majority of these efforts are based on graph classification tasks \citep{r15, r16, r17}, with some limited potency in the field of link prediction \citep{r18, r9}. However, their common feature is their dependence on static networks. Treating dynamic graphs as static graphs can lead to a significant loss of important information. As shown in Figure 1(a), restricting the focus of the meta-learning model only to the graph at time $t_3$ (i.e., static graph works equipped with a meta-learning formulation) is insufficient to capture essential information about the evolution of the network from time $t_1$ to $t_3$, including the time information and the changing preferences of the nodes. These factors have a significant impact on the prediction of dynamic links, especially in dynamic link prediction scenarios for new nodes. 
\begin{figure}
\begin{center}
\includegraphics[width=3.2in]{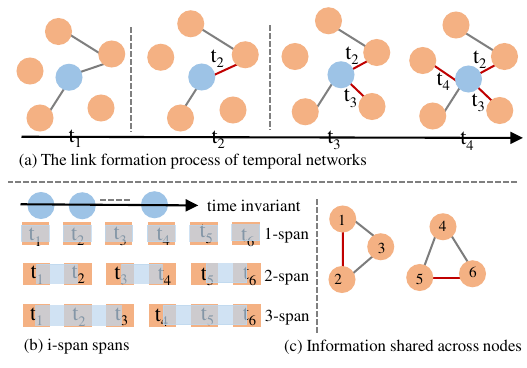}
\caption{(a) The interaction preference changes of nodes during temporal link formation; (b) The time invariance of nodes over different size spans; (c) The shared information across nodes;}
\label{fig1}
\end{center}
\end{figure}

In this paper, we propose DLPNN (\textbf{D}ynamic \textbf{L}ink \textbf{P}rediction for \textbf{N}ew \textbf{N}odes), a novel temporal graph embedding approach endowed with a meta-learning formulation to solve the dynamic link prediction problem for new nodes. To solve the first challenge, we design a temporal graph embedding module with span memory that effectively captures the complex evolutionary properties of nodes. 
The module uses a temporal attention mechanism to generate the node's embedding representation by aggregating the node's intrinsic feature, span dependency, and neighbourhood information. The node pair representations are then fed into a predictor to determine whether temporal links exist.
For the second challenge, we design a meta-learner equipped with a temporal graph embedding module to extract implicit information about two aspects of the dynamic link formation process. The one is the time invariance of node interactions \citep{r29}. As shown in Figure 1(b), we explore the impact of time invariance on the predictive performance of few links by designing spans of different sizes on the support set. 
The other one (Figure 1(c)) we want to extract is information that is shared between nodes \citep{r48}, i.e. nodes that share neighbors are more likely to be connected at a later time.
During the meta-testing phase, the two information obtained by our model can be efficiently adapted to a new node that has sparse interactions for its future links. In summary, our contributions are shown below.
\begin{itemize}
\item{We propose an innovative temporal network embedding approach endowed with meta-learning formulation for dynamic link prediction of new nodes;}

\item{To dynamically obtain the embedding representations of new nodes with few links, we introduce a temporal graph embedding module with node-level span memory. This approach enables the timely and efficient acquisition of embedding representations for new nodes, improving prediction accuracy;}

\item{We devise a meta-learner that extracts insights into time-invariant preferences of node and information shared between nodes during dynamic link formation. This information facilitates rapid adaptation to new nodes in the meta-dynamic link testing phase, enhancing the accuracy of link prediction;}

\item{Experimental results on three publicly available dynamic network datasets show that our proposed method achieves superior results compared to baselines.}
\end{itemize}

\section{Related work}
In this section, we present the related works on dynamic link prediction from two aspects: 1) meta-learning methods that can handle small sample problems, and 2) methods for modeling dynamic graphs. 

\textbf{Meta-learning methods:}
Meta-learning \citep{r14,r20,r21}, also referred to as learning-to-learn, aims to develop models capable of rapid acquisition of new skills or adaptation to novel tasks with minimal training data. MAML \citep{r22} is considered a masterpiece in the realm of meta-learning for its ability to effectively initialize the parameters of a neural network by seamlessly transitioning between various tasks during the training process. Jatin \citep{r17} proposes a meta-learning based few-shot learning method for graph classification, which employs spectral measures of a graph to create super-classes and a super-graph. This approach facilitates improved learning of the underlying relationships between classes. Meta-Graph \citep{r18}, a fusion of meta-learning and GCN \citep{r23}, is employed to predict missing edges in underrepresented samples across multiple graphs. However, it falls short in addressing the link prediction problem within a single graph. G-Meta \citep{r19} is a highly effective approach for both classification and link prediction tasks, leveraging local subgraphs to transfer subgraph-specific information and rapidly acquire transferable knowledge using meta-gradients.
These works are tailored for stationary networks and exhibit commendable performance solely under time-invariant circumstances; However, their performance are greatly compromised in dynamic networks that possess time-varying features. MetaDyGNN \citep{r29}, purports a learning mechanism that involves both interval-wise adaptation and node-wise adaptation, yielding promising results. However, it resorts to neighborhood information aggregation in cases where the number of a node's neighbors exceeds a predetermined threshold, a practice that we deem to be an aberration. Furthermore, randomly sampled time edges have the potential to destroy the evolutionary properties of dynamic graphs. 

\textbf{Dynamic graph models:} 
Currently, there exist two predominant categories of dynamic graph models. The first leverages discrete graph snapshots to represent temporal changes, whereas the second approach characterizes the graph's evolving edges through a sequence of events. EvolveGCN \citep{r30} a discrete method for the tasks of link prediction, node classification and edge classification integrates RNN into GCN and can capture the topological and time information in the same layer. Other similar architectures such as \citep{r31,r33}.
GCRN-M1 \citep{r34} utilizes the state-of-the-art GCN technique to encode each snapshot, followed by leveraging LSTM to capture the temporal dependencies among them. 

For continuous-time dynamic networks, conventional shallow embedding models employ sophisticated techniques such as the Hawkes \citep{r38} process and temporal random walks \citep{r39,r40} to fully capture the intricate process of graph evolution. JODIE \citep{r41} and DyRep \citep{r42} introduced recurrent neural networks (RNN) and its variants to learn and effectively capture the evolutionary process of networks, thereby capturing prolonged dependencies among nodes. Moreover, there exist networks that amalgamate the attention mechanism and RNNs, which excel at capturing not only the fine-grained evolution processes and temporal information but also retain long dependencies of nodes. 
TGAT \citep{r43} combines the unique strengths of GraphSAGE and GAT, thus extending the multi-head attention to temporal graphs and achieving superior results. TGN \citep{r44} elevated the performance and efficiency of the TGAT network by fine-tuning the optimization and incorporating memory modules for nodes. Although these methods have shown remarkable efficacy in addressing dynamic graphs in a broad context, none of them are specifically designed for the intricacies of sparse dynamic graphs. Hence, they are not directly applicable for the task of predicting nodes with few links.

\section{Preliminaries}
\textbf{Temporal graphs:} Formally, a temporal graph can be represented as $G_T=(V_T, E_T)$, where $T$ denotes the survival period of the network. $V_T$ and $E_T$ denote the nodes and temporal edges, respectively. Each edge with timestamp $t$ between node $v$ and $j$ can be denoted by $e_{v,j}(t)=(v, j, f_{v,j}(t), t)$, where $t \in T $, and $f_{v,j}$ represents the feature of the edge. Since they are timestamped edges, there may be multiple edges between the same pair of nodes.

\textbf{New nodes and dynamic link prediction:} We begin by introducing the concept of new nodes. Given an arbitrary time $t$ and its temporal graph $G_t=(V_t, E_t)$, any newly appearing node at the future that has not been previously observed in $G_t$ is defined as a new node.
Next, we introduce the concept of dynamic link prediction for new nodes in the temporal graph. Given a temporal graph $G_t=(V_t, E_t)$ at time $t$, we define $V_{new}$ as the set of new nodes that materialize after $t$. The intention of dynamic link prediction is to accurately predict the future links of each new node $v \in V_{new}$ using its first $N$ interactions. To initiate the construction of the few prediction scenario, a limited number $N$ of historical interactions is selected for each new node within the existing dataset.

\begin{figure}
\begin{center}
\includegraphics[width=3.5in]{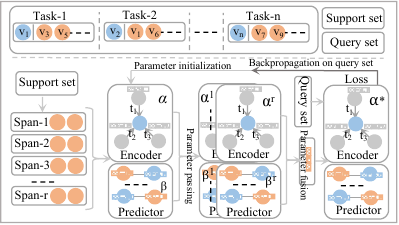} 
\caption{The overall architecture of DLPNN. At the top of the framework shows the process of dividing the temporal network into node tasks and obtaining the support and query sets for each task. At the bottom, the left-hand side shows the division of each node task into training spans. on the right-hand side, information on the two implicit aspects of the link formation process of learning the temporal network by feeding these node tasks and the corresponding spans to the encoder and predictor, as well as the parameter optimisation process of the model.}
\label{fig2}
\end{center}
\end{figure}

\section{Methodology}
In this section, we propose a novel approach, DLPNN, to address the challenges of dynamic link prediction for new nodes. The DLPNN model consists of two components. First, a temporal graph embedding module is designed to dynamically obtain the embedding representations of new nodes, and then these representations are used for dynamic link prediction; Secondly, we design a meta-learner based on the temporal graph embedding module to extract two types of implicit information during the temporal network formation process.

\subsection{Task formulation}
In the meta-learning framework, to extract two potential information of the dynamic link formation process, we view each task as the temporal interaction learning of a single node (For each task, we further divide it into a support set and a test set, as shown in Figure 2(top)). Specifically, we consider the interactions in each task as a temporal binary classification problem between positive and negative links of nodes.
To capture the shared information between nodes, we divide the training set into node-level tasks to learn node-wise information. To extract time-invariant information during dynamic link formation, the support set of each train task should be divided into equally spaced and consecutive interaction spans, while the size of the spans should also be taken into account. This enables a thorough exploration of the effects of distinct size spans on dynamic link prediction. For simplicity, we denote the support set of node $v$ as $C_{v}=\{e_{v,1}(t_1), e_{v,2}(t_2),..., e_{v, N}(t_N)\}$. As shown in Figure 1(b), the time-invariant information under varying size spans are exhibited (e.g., we use $i$-$span$ to represent the size of spans in the support set of node $v$, $i \in [1,2,3,...]$), where the maximum size of spans can be determined based on the support length of node $v$. Formally, these $i$-$span$ spans can be expressed as $S_{v} = \{S_{v}^1: [S_{v,1}^1, S_{v,2}^1,...], S_{v}^2: [S_{v,1}^2, S_{v,2}^2,...],..., S_{v}^i: [S_{v,1}^i, S_{v,2}^i,...] \} = \{S_{v}^1:[[e_{v,1}(t_1)], [e_{v,2}(t_2)], ...], S_{v}^2:[[e_{v,1}(t_1), e_{v,2}(t_2)], [e_{v,3}(t_3), e_{v,4}(t_4)],...], S_{v}^i:[[e_{v,1}(t_1),..., e_{v,i}(t_i)], [e_{v,i+1}(t_{i+1}), e_{v,2i}(t_{2i})],...]..., \}$. Here, it is noted that  redundant interactions within a support set which do not form a complete span are deemed irrelevant and thus excluded. After obtaining these different sizes of spans, we use one size at a time to explore the effect of time-invariant information on prediction performance, and thus determine the optimal span size. Thus, the support set for each task during model training can be represented as $\{S_{v}^i: [S_{v,1}^i, S_{v,2}^i,...]\}$. In addition, we need to select $M$ events that occur after the support set as the query set $Q_{v}=\{e_{v, N+1}(t_{N+1}), e_{v, N+2}(t_{N+2}),..., e_{v, N+M}(t_{N+M})\}$.

In all our experiments, we sample a small number of consecutive interactions for each node to form our training instances on the existing dataset, depending on the direction of network evolution, during the meta-training process.
Specifically, $N$ links are extracted as positive samples and an equal number of negative samples through random negative sampling strategy to form a support set. For the query set, $M$ positive samples and the same number of negative samples are chosen after the support set time. In meta-testing, the $N$ dynamic links of each new node serve as a fine-tuning support set, while all remaining links are used as query set. It is crucial to note that, in both the support and query sets, each positive sample occurrence should be continuous based on the previous interaction to comply with the network's natural evolutionary properties, rather than randomly sampled from all interactions of the node (i.e., jumping is not allowed).

\subsection{Temporal graph embedding for dynamic link prediction} 
In this subsection, we present an encoder $f_{\alpha}$ for the dynamic node embedding, and a predictor, $f_{\beta}$ for the dynamic link prediction. $\alpha$ and $\beta$ denote the learnable parameters in the encoder and predictor. The embedding representation of node $v$ at time $t$ can be denoted as $z_v(t)=f_{\alpha}(v, t, \mathcal{N}_{v}(t))$, which aggregates information from itself, span memory, and its temporal neighbors. Specifically,  the encoder inputs include, $h_{v}=W_s x_{v} + m_{v}^{r-1}$, $\Phi(\Delta t)$, and $h_{j}=W_s x_{j}, j \in \mathcal{N}_{t}(v)$. $x_v$ denotes the static feature of node $j$. $\Phi(\Delta t)$ represents the time encoding. $W_s$ denotes the linear transformation parameter and $m_{v}^{r-1}$ denotes the span memory of node $v$. Since this is a node-level task, node $j$ is only linearly transformed.
The predictor is designed to determine the likelihood of the current link $e_{v,j}(t)$ occurring. Subsequently, we will provide a detailed description of each of them.

\textbf{Encoder:} Here, we propose using a temporal attention approach with span memory  to effectively aggregate the neighborhood information. 
Precisely, we utilize node $v$ and its neighbors to construct the query information $q_{v}(t) = W_q [h_{v} \left |  \right | \Phi(0)]$, key information $k_{j}(t) = W_k [h_{j} \left |  \right | \Phi(t-t_{j}) \left |  \right | f_{v,j}(t_j)]$ respectively. Where $\left |  \right |$ is the concatenation operator, and $W_q$, $W_k$, $W_v$ and $W_{ts_1}$, $W_{ts_2}$ denote the parameter matrices. Next, attention weight $ws$ is computed for each neighbor using the query and key information. The neighborhood information $h_{\mathcal{N}_{t}(v)}$ of node $v$ is then aggregated based on these weights:

\begin{equation}
\label{eq:eq1}
\begin{aligned}
ws =softmax(\frac{q_v(t) k_j(t)^{T} }{\sqrt{d_k}})  \quad j \in \mathcal{N}_{t}(v),
\end{aligned}
\end{equation}

\begin{equation}
\label{eq:eq2}
\begin{aligned}
h_{\mathcal{N}_{t}(v)} = \sum_{j \in \mathcal{N}_{t}(v)} ws\cdot W_v [h_{j} \left |  \right | \Phi(t-t_{j}) \left |  \right | f_{v,j}(t_j)] , 
\end{aligned}
\end{equation}

\begin{equation}
\label{eq:eq3}
\begin{aligned}
h_{\mathcal{N}_{t}(v)} = W_{ts_1}(W_{ts_2}  h_{\mathcal{N}_{t}(v)} \left |  \right | q_v(t)), 
\end{aligned}
\end{equation}
where $d_k$ is the scaling factor. $\Phi(\Delta t)$  represents a time encoding function that describes the temporal pattern of network evolution with explicit time features. In this study, random Fourier features are employed as our method for time encoding:
\begin{equation}
\label{eq:eq4}
\begin{aligned}
\Phi(\Delta t) = & \sqrt{\frac{1}{d}}[cos(\lambda_1 \Delta t), sin(\eta_1 \Delta t),...,\\
               & cos(\lambda_d \Delta t), sin(\eta_d \Delta t)],
\end{aligned}
\end{equation}
where ${\lambda}$ and ${\eta}$ are parameters that can be trained, and $d$ denotes the dimension of the time embedding vector. 
After acquiring the neighborhood aggregation information, we proceed to compute the embedding for node $v$,

\begin{equation}
\label{eq:eq5}
\begin{aligned}
z_{v}(t) = \delta ((h_{\mathcal{N}_{t}}(t) \left |  \right | h_{v}) W_o),
\end{aligned}
\end{equation}
where $\delta$ and $W_o$ denote the activation function and fusion parameters, respectively. Our embedding method is also applicable to the aggregation of temporal multi-hop neighbors.


To mitigate the inherent challenge posed by the scarcity of interactions for new nodes, we devise a  node-level span memory to effectively capture the span dependency. Specifically, after processing each span $S_{v,r}^i$ ($S_r^i \in \{S_{v}^i: [S_{v,1}^i, S_{v,2}^i,...,S_{v,r}^i,...]\}$), we use the latest interaction information for node $v$ in that span as our span information $I_{S_{v,r}^i}$. It is expressed as $[z_{v}(t)\left | \right | z_{j}(t)\left | \right | f_{v,j}(t) \left | \right | \Phi(t-t^-)]$. where, $t^-$ represents node $v's$ most recent interaction at time $t$. Here we use the connection method to obtain the interval information, but other methods such as linear aggregation are also possible. After obtaining the span information, we update it into the span dependency, $m_v^{r-1}$:
\begin{equation}
\label{eq:eq6}
\begin{aligned}
m_v^r = \mathcal{U}(m_v^{r-1}, I_{S_{v,r}^i}),
\end{aligned}
\end{equation}

where $m_v^r$ is the span dependency of node $v$, i.e. the dependency from span $1$ to $r$. $\mathcal{U}(.)$ represents the update function which, in this case, is implemented using the GRU function.
The span dependency we obtain will be used in the calculation of the next span.

\textbf{Predictor:}
Dynamic link prediction aims to determine whether there is a temporal edge between two nodes, making it a binary classification problem. To achieve this goal, we employ the MLP architecture as our predictor. The encoder generates node embeddings for $v$ and $j$ at  time $t$. We then utilize the predictor to classify whether the temporal edge $e_{v,j}(t)$ exists or not.

\begin{equation}
\label{eq:eq7}
\begin{aligned}
p(e_{v,j}(t))&=f_{\beta}(z_v(t), z_j(t))\\
&=sigmoid(MLP(z_v(t) \left |  \right | z_j(t)))
\end{aligned}
\end{equation}
 where the MLP represents a multilayer perceptron with two layers. 

\subsection{Meta-learner for extracting implicit information}
In this section, we use an adaptive meta-learner to extract two types of implicit information (As shown at the bottom of Figure 2). At the first level of adaptation, we fine-tune the parameters for both the encoder and predictor using separate learning rates at each span. As shown in subsection B, the support set is composed of equally spaced interaction fragments, represented by $\{S_{v}^i: [S_{v,1}^i, S_{v,2}^i,...S_{v,r}^i,...]\}$. The span adaptation updates the parameters based on the loss of each span $S_{v,r}^i$ in  $S_{v}^i$, to capture node $v$'s time invariance in the $r$-th span and improve the model's accuracy.

In particular, for each temporal edge $e_{v,j}(t^r)=(v,j, f_{v,j}(t^r), t^r)$ in the $r$-th span $S_{v,r}^i$ of the support set, the computation of the embedding representation for nodes $v,j$,  along with the probability for edge $e_{v,j}(t^r)$ can be achieved sequentially by following steps:
\begin{equation}
\label{eq:eq8}
\begin{aligned}
z_{v}(t^r) = f_{\alpha}(v, t^r, \mathcal{N}_{v}(t^r)),
\end{aligned}
\end{equation}

\begin{equation}
\label{eq:eq9}
\begin{aligned}
z_{j}(t^r) = f_{\alpha}(j, t^r, \mathcal{N}_{j}(t^r)),
\end{aligned}
\end{equation}

\begin{equation}
\label{eq:eq10}
\begin{aligned}
p(e_{v,j}(t^r))=f_{\beta}(z_{v}(t^r) , z_{j}(t^r)).
\end{aligned}
\end{equation}

We then calculate the span loss based on the positive and negative links of node $v$.
\begin{equation}
\label{eq:eq11}
\begin{aligned}
\mathcal{L}^r = \sum_{e_{v,j}(t^r), j\in \mathcal{N}_{v}(t^r)} log(p(e_{v,j}(t^r)))\\
-\sum_{e_{\Omega}\in \Omega(v)}log(1-p(e_{\Omega})),
\end{aligned}
\end{equation}
where $e_{\Omega}$ denotes the negatively sampled edges and $\Omega(v)$ denotes the negative sampling distribution. To maintain balance during the training process, we equate the number of negative sample edges with that of positive ones.
We then perform global adaptation of the encoder and predictor parameters using gradient descent.
\begin{equation}
\label{eq:eq12}
\begin{aligned}
\alpha_v^r = \alpha -{lr_1} \frac{\partial\mathcal{L}^r(S_{v,r}^i)}{\partial \alpha },
\end{aligned}
\end{equation}

\begin{equation}
\label{eq:eq13}
\begin{aligned}
\beta _v^r = \beta -lr_2 \frac{\partial\mathcal{L}^r(S_{v,r}^i)}{\partial \beta },
\end{aligned}
\end{equation}
where $lr_1$ and $lr_2$ are the learning rates of the encoder and predictor, respectively. 

After adapting for time invariance, we assign node-specific information to the predictor to address task-specific requirements (i.e. the node-level adaptation). More prescisely, we perform parameter re-tuning on the predictor to further enhance the predictive capacity of the model for new nodes. The adjusted parameters contain node-specific information tailored to the requirements of the current task.  Through this approach, we achieve node-level adaptation by performing the following calculation:
\begin{equation}
\label{eq:eq14}
\begin{aligned}
\beta_v^r = W_{s} x_{v} +\beta_v^r.
\end{aligned}
\end{equation}

Drawing on our previous steps, we have implemented meta-learning to adapt the global parameters of each span from $(\alpha_v, \beta_v)$ to $(\alpha_v^r, \beta_v^r)$. In the subsequent meta-test phase, these parameters from different spans are fused together using differentiated weights $w_{v}^{r}$, thereby enabling enhanced predictive accuracy and performance:
\begin{equation}
\label{eq:eq15}
\begin{aligned}
w_{v}^{r} = \frac{exp(-\mathcal{L}(\alpha_v^{r}, \beta_v^{r}, Q_{v})}{ {\textstyle \sum_{r=1}^{num(r)} exp(-\mathcal{L}(\alpha_v^{r}, \beta_v^{r}, Q_{v}))}},
\end{aligned}
\end{equation}

\begin{equation}
\label{eq:eq16}
\begin{aligned}
\bar{\alpha_v}=\sum_{r=1}^{num(r)} w_{v}^{r} \alpha_v^{r},
\end{aligned}
\end{equation}

\begin{equation}
\label{eq:eq17}
\begin{aligned}
\bar{\beta_v}=\sum_{r=1}^{num(r)} w_{v}^{r} \beta_v^{r},
\end{aligned}
\end{equation}

where, $num(r)$ represents the total number of spans, while $\bar{\alpha_v}$ and $\bar{\beta_v}$ refer to the fused parameters. Subsequently, leveraging these adapted parameters $(\bar{\alpha_v}, \bar{\beta_v})$, we calculate the model's loss on set using backpropagation
We use these merged parameters to compute the loss of the model on the support set, $Q_{v}$  of node $v$, and then update the DLPNN parameters by a back-propagation algorithm:
\begin{equation}
\label{eq:eq18}
\begin{aligned}
\theta \leftarrow \theta - lr_3\nabla_\theta \sum_{v\in V_T} \mathcal{L}(\bar{\alpha_v}, \bar{\beta_v}, Q_{v}),
\end{aligned}
\end{equation}
where $lr_3$ denotes the overall meta-learning rate, and $ \theta$ denotes all learnable parameters in the model. The learning process of TLPNN is described in Algorithm 1. The validity of the model will be analysed in the experiments.

\begin{algorithm}[t]
\caption{The learning process of TLPNN.}
\label{alg:TLPNN}
\textbf{Require}: Temporal graph $G_T$, hyperparameters, Initialisation of model parameters, $\theta$. \\
\textbf{Data preparation}: For each node $v$, a node-level task is constructed, and each task is divided into a support set, $\{S_{v}^i: [S_{v,1}^i, S_{v,2}^i,...]\}$ and a validation set, $Q_v$.

\begin{algorithmic}[1] 
\FOR{each node $v$ in the batch} 
\FOR{all dynamic links in span $S_{v,r}^i \in S_{v}^i$}
\STATE{calculate $\mathcal{L}^r(S_{v,r}^i)$ to perform the two adaptions. \hfill $\triangleright$ See~ \eqref{eq:eq11}}
\STATE {span adaption: 
$\alpha_v^r \leftarrow \alpha -{lr_1} \frac{\partial\mathcal{L}^r(S_{v,r}^i)}{\partial \alpha }$, 
$\beta _v^r \leftarrow \beta -lr_2 \frac{\partial\mathcal{L}^r(S_{v,r}^i)}{\partial \beta }$. 
\hfill $\triangleright$ 
See~\eqref{eq:eq12} and \eqref{eq:eq13}.} 
\STATE{ node adaption: $\beta_v^r \leftarrow W_{s} x_v +\beta_v^r$. \hfill $\triangleright$ See~\eqref{eq:eq14}.}

\ENDFOR
\STATE{The fusion of parameters for each span using a weighting mechanism: $\bar{\alpha_v}\leftarrow\sum_{r=1}^{num(r)} w_{v}^{r} \alpha_v^{r}$, $\bar{\beta_v}\leftarrow\sum_{r=1}^{num(r)} w_{v}^{r} \beta_v^{r}$.  \hfill $\triangleright$ See~\eqref{eq:eq15}, \eqref{eq:eq16} and \eqref{eq:eq17}}
\ENDFOR
\STATE{Back-propagate the global parameter,$\theta$. \hfill $\triangleright$ See~\eqref{eq:eq18} }
\end{algorithmic}
\end{algorithm}

\section{Experiments}
We evaluate the proposed method on three public dynamic datasets, i.e., Wikipedia, DBLP and Reddit. We compare DLPNN with three types of benchmarks on the sparse link prediction task and report experimental results to answer four research questions. RQ1: How does DLPNN perform in comparison with the state-of-the-art works?
RQ2: To what degree do hyperparameters impact the performance of DLPNN? RQ3: How significantly does DLPNN profit from the utilization of meta-learning and hierarchical adaptation techniques? RQ4: How time efficient is DLPNN? and how effective is multi-step adaptation?

\subsection{Dynamic datasets}
\textbf{Wikipedia} \citep{r42}: This dataset was collected over a one-month timespan from the Wikipedia platform. It includes the top 1000 most edited pages and the 9227 most active editors, resulting in a total of 157475 timestamped edges. Each of these temporal edges has been transformed into a LIWC feature vector.

\textbf{Reddit} \citep{r42}: The Reddit dataset consists of the number of postings made by users on the Reddit website over the course of a month. It encompasses 10984 nodes and 672448 temporal edges. Similar to the Wikipedia dataset, its edges have also been converted into LIWC feature vectors.

\textbf{DBLP} \citep{r47}: The DBLP dataset comprises academic citation records and encompasses 28085 author nodes. These authors have the ability to collaborate, resulting in 286894 temporal edges occurring at specific times.

\subsection{Baselines}
We select static graph methods, dynamic network models, and meta-based graph technologies as our baselines.

\textbf{Static methods:} GraphSAGE \citep{r45} and GAT \citep{r46} are local inductive aggregation methods. GraphSAGE comprises of neighbor sampling and neighborhood aggregation, while GAT captures the importance of different neighbors through multi-head attention for node embedding. Both of these methods can be easily adapted to dynamic link prediction tasks. Two-layer GraphSAGE and GAT were leveraged in our experiments and followed the other settings in the original paper.

\textbf{Dynamic methods:} We choose DyRep \citep{r42}, EvolveGCN \citep{r30} and TGAT \citep{r43} as our dynamic baselines. DyRep is typical model based on temporal point process, and in our experiment, we only use the core node embedding moudle for dynamic link prediction task. EvolveGCN first obtains the node representation on each snapshot using the GCN, and then operates RNNs over snapshots. TGAT, a continuous-time dynmaic method, utilizes temporal attention mechanism to aggregate the neighbors of a node.

\textbf{Meta-based methods:} For this component, we select two combinatorial algorithms, as well as two existing meta-learning models. Specifically, GS-MAML and T-MAML simply incorporate GraphSGAE \citep{r45} and TGAT \citep{r43}, respectively, into a meta-learning framework \citep{r22} to establish our benchmark models. Meta-GCN \citep{r15} has been selected for its ability to effectively tackle the challenge of few-shot node classification, with its network architecture and loss computation being well-suited for dynamic link prediction requirements. MetaDyGNN \citep{r29} integrates meta-learning into dynamic networks to address the challenge of small-sample link prediction. The initial implementation of this method involved the random sampling of data from a continuous stream of interactions, which would have led to a loss of crucial information in the dynamic network. In our experimental setup, we implemented the optimal configuration from the original paper, ensuring a rigorous and reliable evaluation of the method's efficacy.

\begin{table}[!h]
    \caption{The results of dynamic link prediction for new nodes with ACC(\%), Macro-F1(\%) and the area under the ROC curve (AUC(\%)) on Wikipedia dataset. The best results on each dataset are bolded.}
    \begin{center}
    \setlength{\tabcolsep}{5.7mm}{
    \begin{tabular}{llll}  
    \toprule[1.0pt]
    \textbf{}          & \textbf{Macro-F1} & \textbf{ACC} & \textbf{AUC} \\ 
    \hline
    \textbf{Wikipedia} &                   &              &              \\
    GAT                & 74.94             & 76.03        & 78.76        \\
    GraphSAGE          & 74.76             & 75.84        & 77.17        \\
    EvoleGCN           & 54.27             & 56.36        & 63.15        \\
    DyRep              & 56.93             & 58.01        & 62.54        \\
    TGAT               & 86.89             & 87.15        & 90.44        \\
    GS-MAML           & 77.49             & 79.54        & 81.21        \\
    T-MAML           & 83.04             & 83.29        & 85.04        \\
    Meat-GNN           & 77.13             & 78.92        & 80.96        \\
    Meta-DyGNN          & 91.63             & 89.67        & 94.10        \\
    DLPNN              & \textbf{93.14}    & \textbf{91.37 }   & \textbf{95.24 }       \\ 
    \bottomrule[1.0pt]
    \end{tabular} }
    \end{center}
    \label{tab:t2}
\end{table}

\begin{table}[htb]
\caption{The dynamic link prediction results on Reddit dataset (AUC(\%), ACC(\%)), Macro-F1(\%)).}
\begin{center}
\setlength{\tabcolsep}{5.7mm}{
\begin{tabular}{llll}
\toprule[1.0pt]
           & \textbf{Macro-F1} & \textbf{ACC }  & \textbf{AUC}   \\ \hline
\textbf{Reddit }    &          &       &       \\
GAT        & 88.64    & 89.86 & 93.36 \\
GraphSAGE   & 88.28    & 89.32 & 93.21 \\
EvoleGCN    & 57.13    & 58.21 & 62.64 \\
DyRep        & 61.27    & 62.14 & 66.08 \\
TGAT         & 93.27    & 93.53 & 95.45 \\
GS-MAML     & 85.91    & 87.41 & 91.34 \\
T-MAML       & 87.09    & 87.46 & 91.07 \\
Meat-GNN   & 85.44    & 86.27 & 90.87 \\
Meta-DyGNN  & 92.96    & 93.09 & 96.49 \\
DLPNN      & \textbf{94.47}    & \textbf{94.30} & \textbf{98.10} \\
\bottomrule[1.0pt]
\end{tabular} }
\end{center}
\label{tab:t3}
\end{table}

\begin{table}[htb]
\caption{The dynamic link prediction results on DBLP dataset (AUC(\%), ACC(\%)), Macro-F1(\%)).}
\begin{center}
\setlength{\tabcolsep}{5.7mm}{
\begin{tabular}{llll}
\toprule[1.0pt]
 \textbf{}             & \textbf{Macro-F1} & \textbf{ACC}   & \textbf{AUC}   \\ \hline
\textbf{DBLP} &                   &                &                \\
GAT            & 72.56             & 73.18          & 77.16          \\
GraphSAGE      & 71.12             & 72.17          & 76.35          \\
EvoleGCN       & 56.92             & 57.98          & 64.14          \\
DyRep          & 57.79             & 59.14          & 66.73          \\
TGAT           & 76.26             & 76.32          & 81.13          \\
GS-MAML       & 75.12             & 76.18          & 80.17          \\
T-MAML        & 72.67             & 73.86          & 78.03          \\
Meat-GNN      & 74.81             & 75.48          & 79.64          \\
Meta-DyGNN     & 80.84             & 80.31          & 87.33          \\ 
DLPNN         & \textbf{81.00}    & \textbf{80.65} & \textbf{87.40} \\ \bottomrule[1.0pt]
\end{tabular} }
\end{center}
\label{tab:t4}
\end{table}

\begin{table}
\caption{The dynamic link prediction results on three datasets at different N (AUC(\%)).}
\begin{center}
\setlength{\tabcolsep}{5.7mm}{
\begin{tabular}{llll}
\toprule[1.0pt]
             & \textbf{Wikipedia} & \textbf{Reddit} & \textbf{DBLP}  \\ \hline
N=2 & 87.04     & 94.84  & 79.63  \\
N=4 & 93.66     & 96.15  & 82.30  \\
N=6 & 93.78     & 96.09  & 86.00  \\
N=8 & 95.24     & 98.10  & 87.40  \\
\bottomrule[1.0 pt]
\end{tabular}}
\end{center}
\label{tab:t5}
\end{table}

\subsection{Experimental setup}
Our evaluation task is the dynamic link prediction of new nodes that are not observed during the training process. In our experiments, we first divide all temporal edges into three edge sets $D_{train}$, $D_{val}$ and $D_{test}$ in a ratio of 6:2:2 according to the evolution direction of the network. Then, based on the three edge sets, we generate the training, validation and testing sets. Specifically, all the nodes and their associated edges in $D_{train}$ are set as the training set; the validation set consists of the nodes whose first appearance occur in $D_{val}$ and their related edges; as for the testing set, we choose the nodes that only appear after $D_{val}$ and their corresponding edges.  At the validation and test steps, we assume that each node has the first known $N$ edges (2, 4, 6, 8). Then, given the same amount of randomly picked edges, the goal is to determine whether a future edge will occur. To fully evaluate all methods, three metrics (i.e., ACC, AUC and Macro-F1) are used in our experiment.

For a fair comparison, all experiments use the same learning rate, batch size, $N$, and embedding dimension as follows. We use the Adma optimizer to optimize our model parameters. We set batch size to 64 and node embedding dimension to 64 for all three datasets. For each node, we sample $N$ neighbors for neighborhood aggregation (e.g., N=8. In the experiment with the hyperparameter $N$, the number of neighbours aggregated is also set to the same values of $N$ as 2, 4, 6, 8). The maximum number of epochs is 30. The learning rates for meta-learner, time span adaption and node adaption are 0.001, 0.0002 and 0.025. Two-hop temporal neighbors and two heads of attention are used in our experiments. 

\subsection{The results of link prediction (RQ1)}
In this subsection, We demonstrate dynamic link prediction results on three datasets and compare our method with three types of baselines. In Table \ref{tab:t2}, \ref{tab:t3} and \ref{tab:t4}, we set $N$=8 for all methods. The best results on each dataset are represented by bold numbers, and we have the following observations.

(1) In general, our proposed algorithm has achieved the best results in Table \ref{tab:t2}, \ref{tab:t3}, \ref{tab:t4}, especially on the Reddit dataset, where all three evaluation matrixs have achieved the highest values by ACC: 94.30\%, AUC: 98.10\% and Macro-F1: 94.47\%. In table \ref{tab:t5}, the performance of our model gradually improves as the $N$ increases. Experimental results in four tables demonstrate the effectiveness of DLPNN on link prediction for new nodes with few links. 

(2)  Compared with the static methods (i.e., GraphSAGE and GAT) and their variants (i.e., Meta-GNN and GraphSAGE-MAML), the method we proposed make great improvements. For example, in Tables \ref{tab:t2}, \ref{tab:t3} and \ref{tab:t4}, the AUC results of our method are 16.48\%, 4.74\% and 10.24\% higher than those of GAT, respectively. The main reason for this phenomenon is that the time information in the dynamic graph is considered in our model. Embedding explicit time information into temporal graph attention enables the model to capture temporal patterns of network evolution. Besides, the meta-adaptive learning over spans also contributes to our model's strong link prediction ability.

\begin{figure*}
\centering
\subfloat[Neighbor\_AUC]{\includegraphics[width=1.7in]{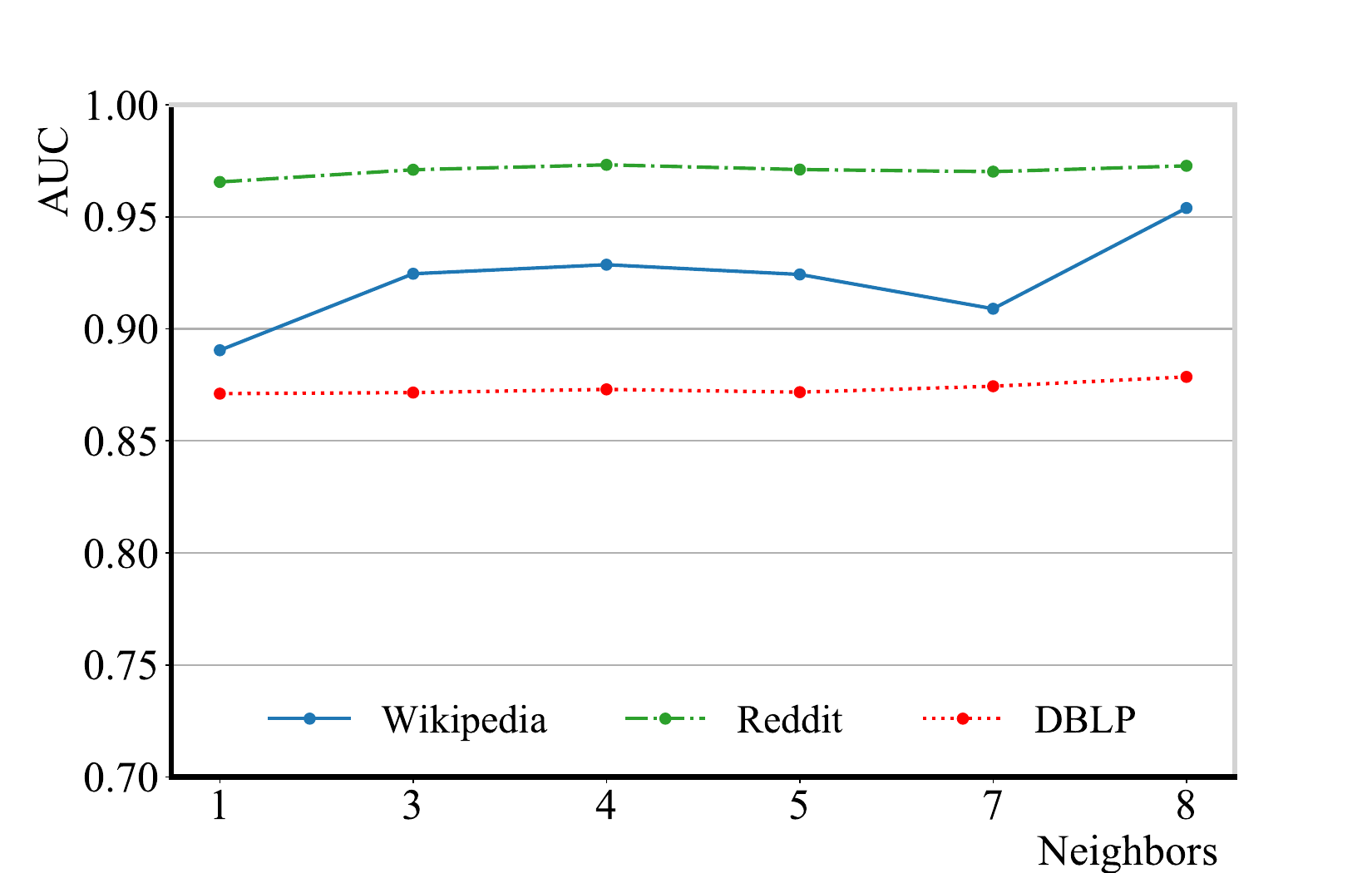}%
\label{fig3-(a1)}}
\hfil
\subfloat[Batchsize\_AUC]{\includegraphics[width=1.7in]{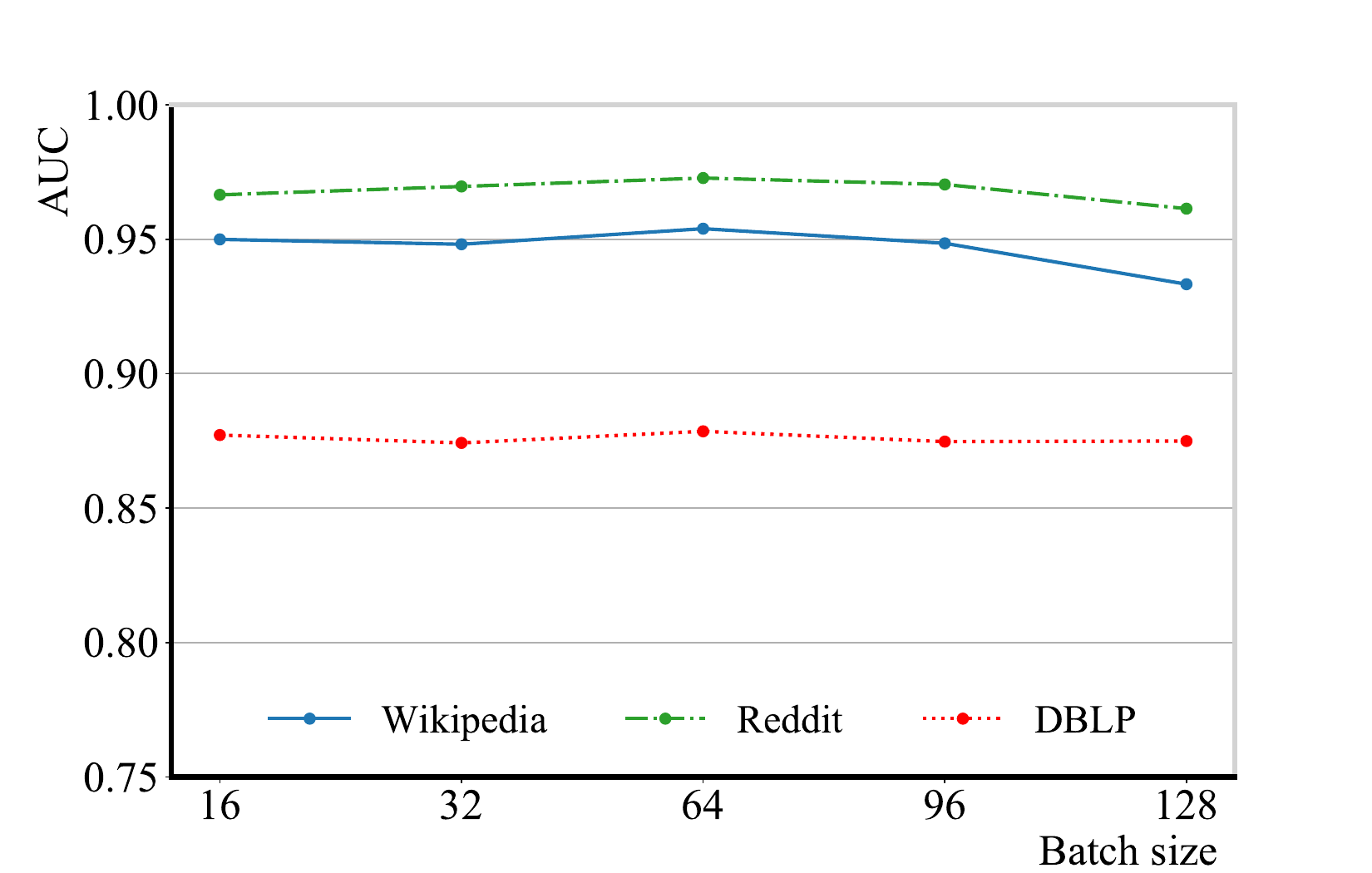}%
\label{fig3-(a2)}}
\hfil
\subfloat[Embedding\_AUC]{\includegraphics[width=1.7in]{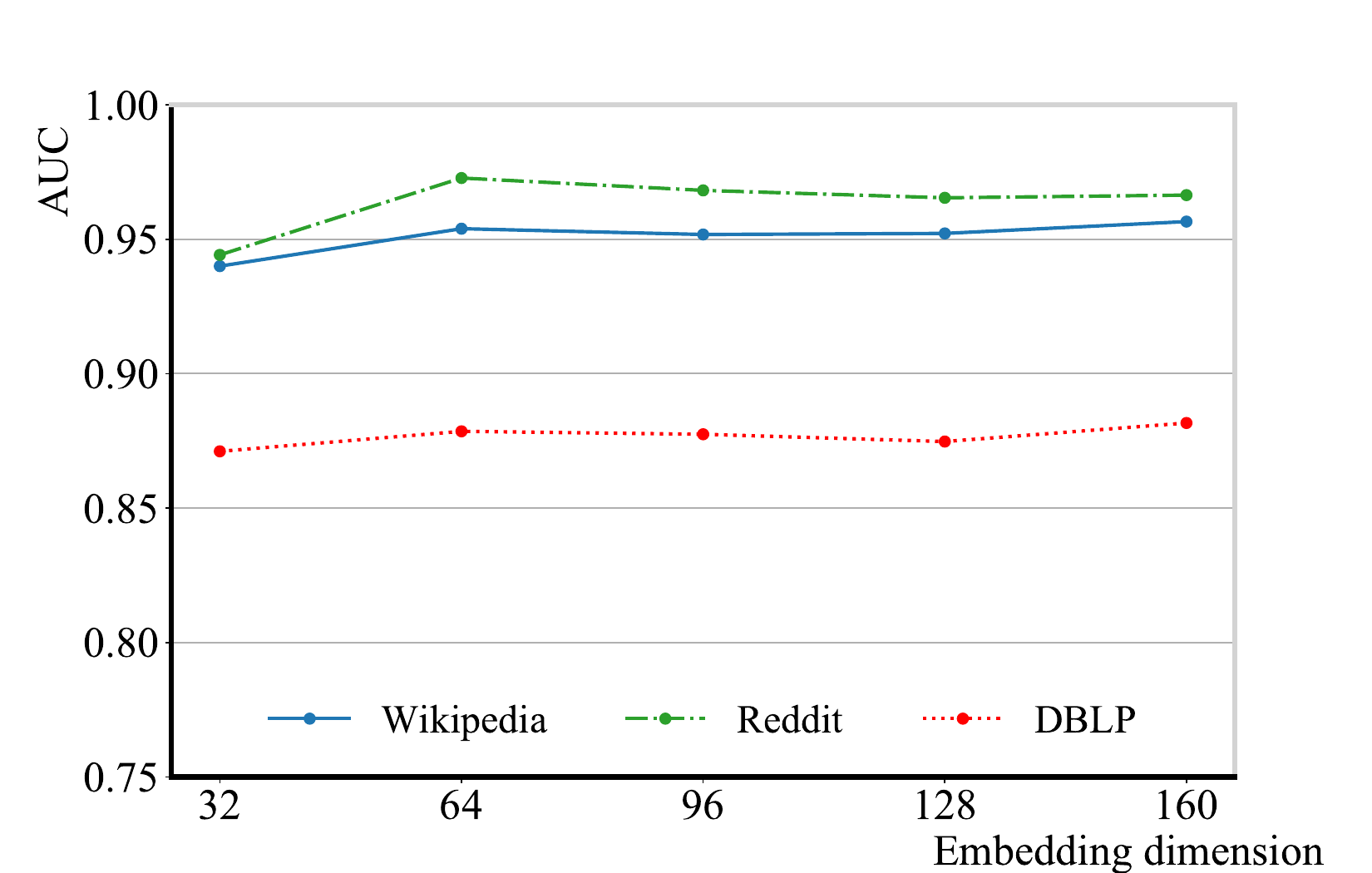}%
\label{fig3-(a3)}}
\hfil
\subfloat[Span\_AUC]{\includegraphics[width=1.7in]{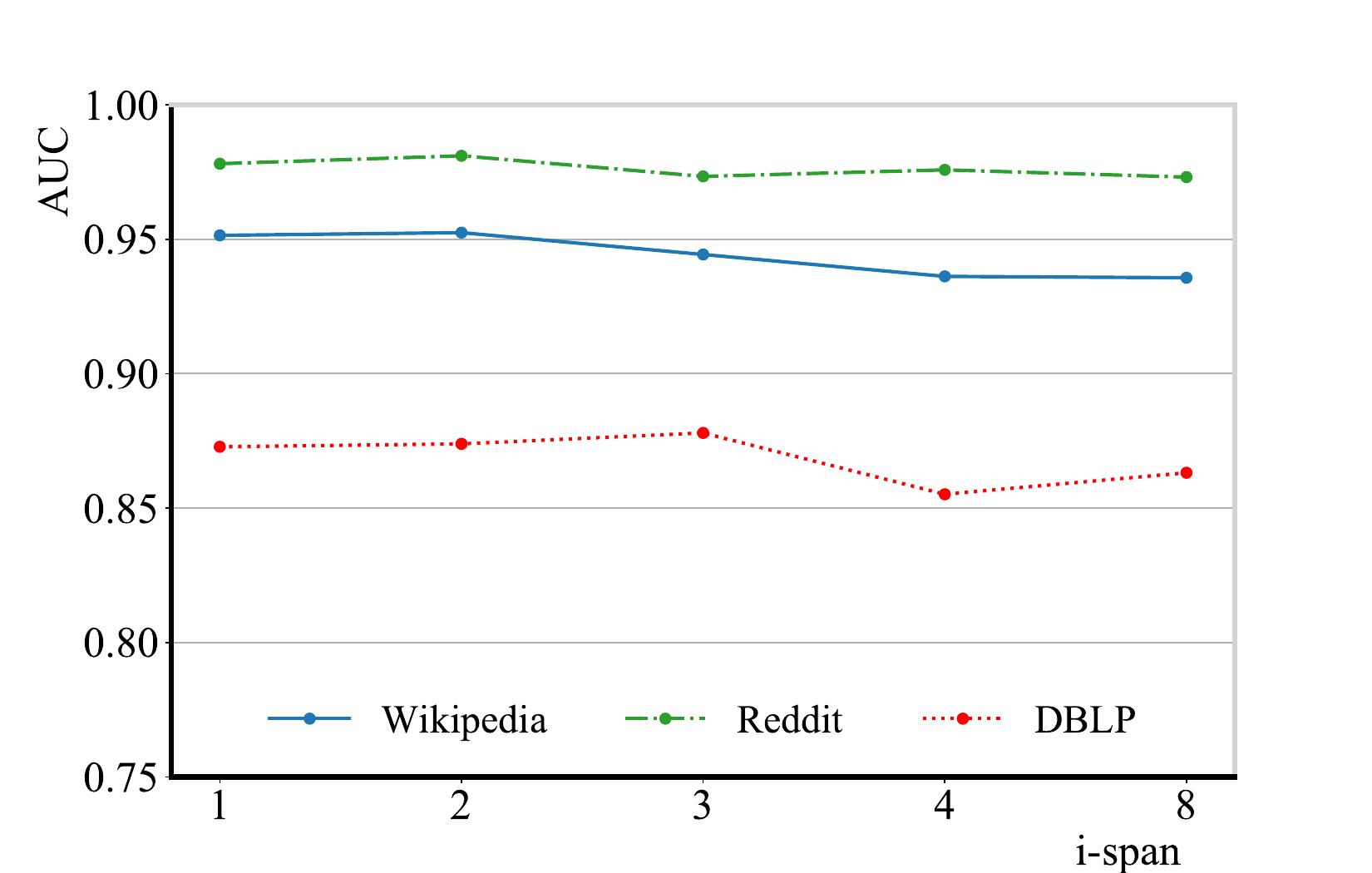}%
\label{fig3-(a4)}}

\caption{The time taken by the different models to achieve the best predictions on the three datasets.}
\label{fig3}
\end{figure*}

\begin{figure}
\centering
\subfloat[AUC]{\includegraphics[width=2.0in]{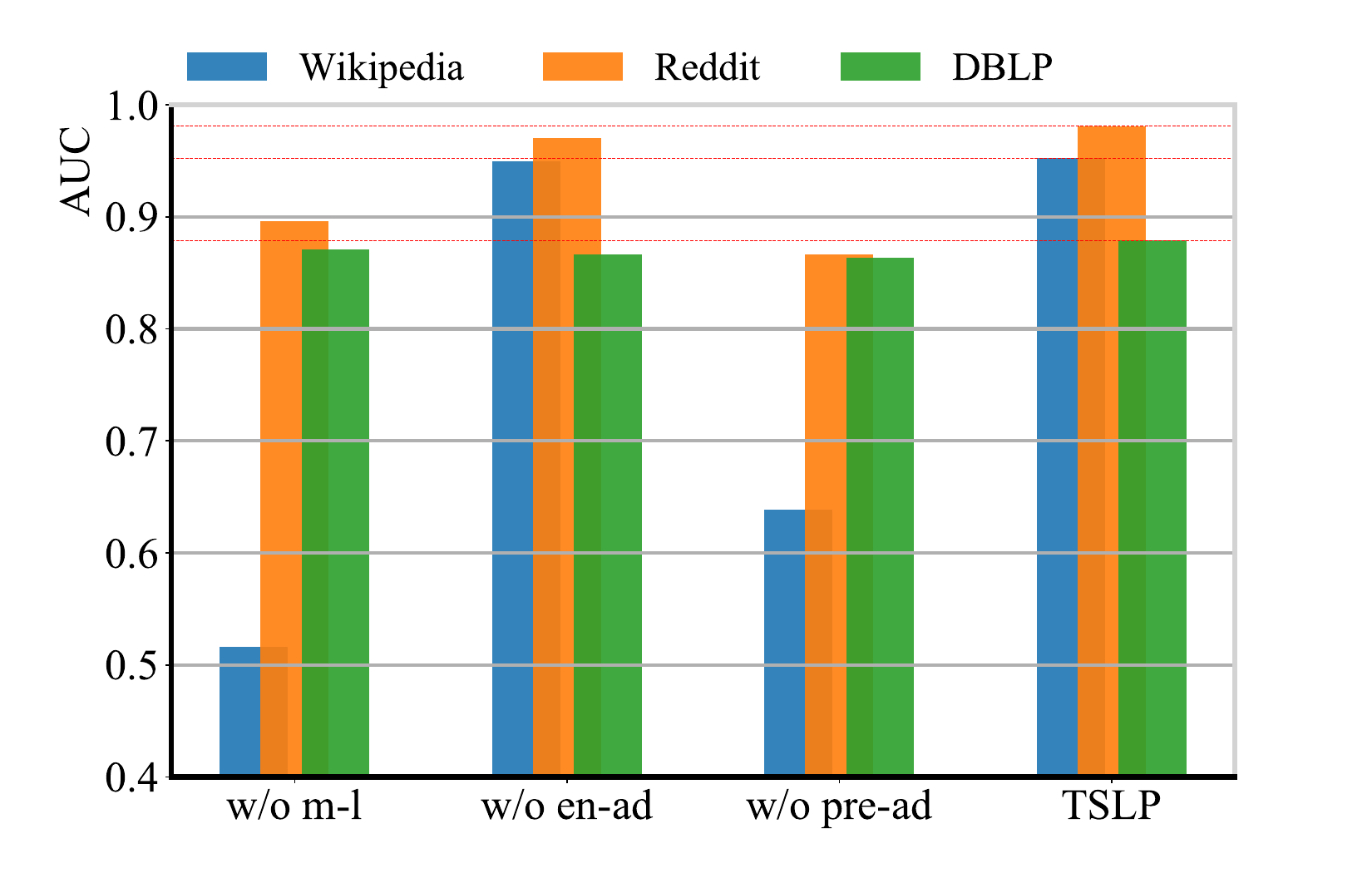}%
\label{fig4-(a1)}}
\caption{The ablation experiments. w/o m-l: DLPNN without meat-learning; w/o en-ad: DLPNN without encoder adapation; w/o pre-ad: DLPNN without predictor adapation.}
\label{fig4}
\end{figure}

(3) For the dynamic models, EvolveGCN, DyRep and TGAT, we observe an interesting phenomenon that EvolveGCN and DyRep perform worse than the static models. For example, the ACC of EvolveGCN on the Reddit dataset in Table \ref{tab:t3} is 19.67\% lower than GAT.  We suggest that the possible reason for this is that their dynamic embedding designs in inherent scenarios cannot adapt to new dynamic scenarios. For their comparison with DLPNN, DLPNN captures the time invariance of nodes and shared information between nodes through meta-learning and can quickly adapt to new nodes. In contrast, the parameters of the three dynamic models after training are generalisations of the large number of historical interactions of the nodes, and they cannot give better predictive performance for new nodes with a small number of historical interactions (e.g., DLPNN achieves higher AUC results than TGAT across various datasets, with a 4.80\% increase on Wikipedia, 2.65\% increase on Reddit, and 6.27\% increase on DBLP).

(4) Compared to TGAT+MAML, our model is much better in terms of prediction performance, e.g. the results on the Wikipedia dataset are about 10\% higher in Table \ref{tab:t2}. In terms of computational resource consumption, our model saves more computational resources due to its proper design. Compared to MetaDyGNN, our model performs better in terms of time efficiency and prediction performance. In addition, MetaDyGNN suffers from two issues. Firstly, the number of aggregated neighbors greatly surpasses the number of historical interactions observed for new nodes,  which is unreasonable. Secondly, the edges of new nodes should not be randomly sampled in the temporal graph, as this fails to account for the natural evolution of the network. These comparisons demonstrate the effectiveness and advantages of DLPNN in dynamic link prediction.

\begin{figure*}
\centering
\subfloat[Wikipedia dataset]{\includegraphics[width=2.0in]{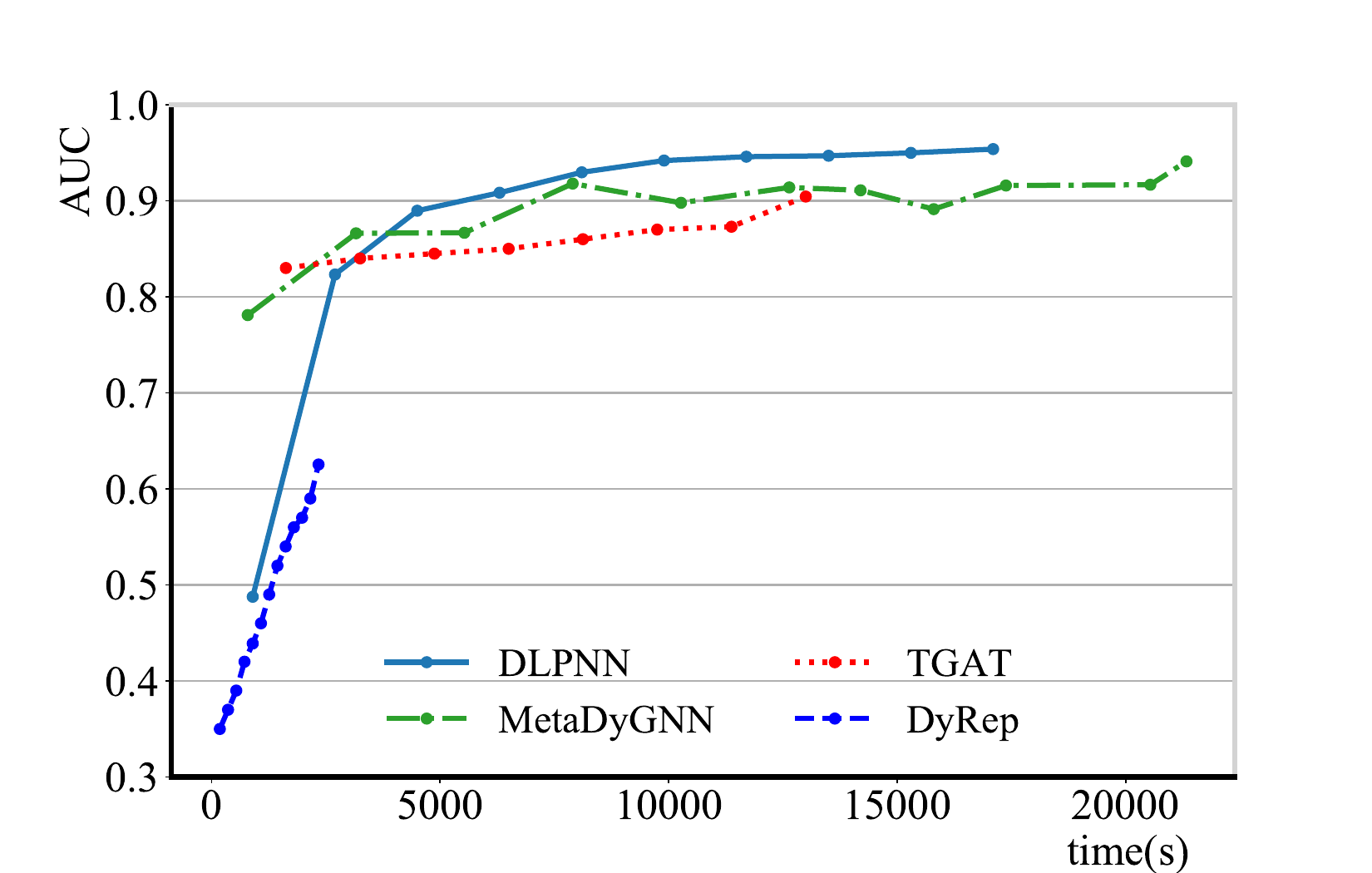}%
\label{fig5-(a1)}}
\hfil
\subfloat[Reddit dataset]{\includegraphics[width=2.0in]{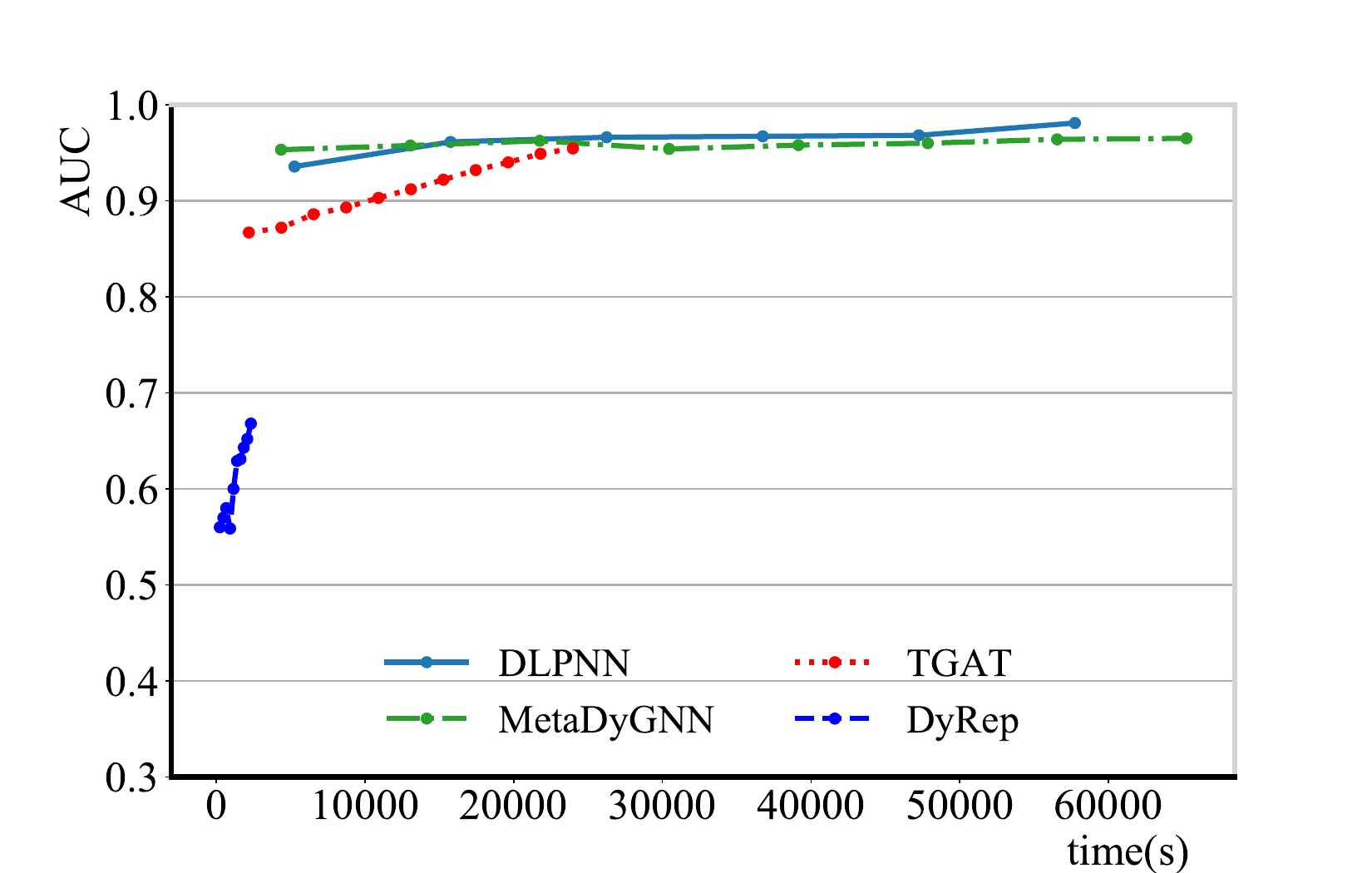}%
\label{fig5-(a2)}}
\hfil
\subfloat[DBLP dataset]{\includegraphics[width=2.0in]{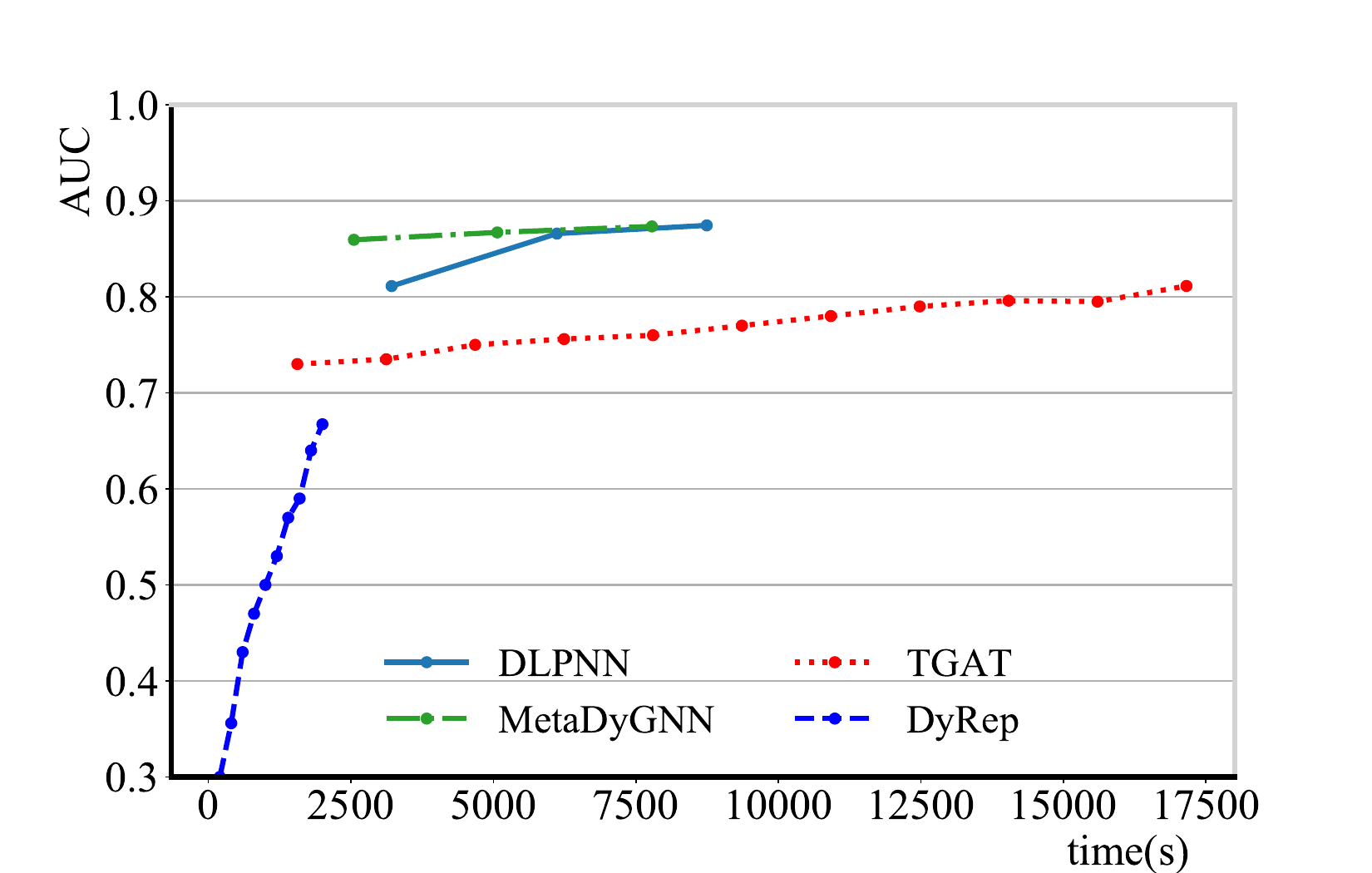}%
\label{fig5-(a3)}}
\caption{The time taken by the different models to achieve the best predictions on the three datasets.}
\label{fig5}
\end{figure*}

\begin{table}[htb]
\caption{The dynamic prediction results of DLPNN with different numbers of adaptation steps on the Wikipedia dataset. Adaption: 0-4 indicates 0 to 4 adaptation steps.}
\begin{center}
\setlength{\tabcolsep}{5.0mm}{
\begin{tabular}{llll}
\toprule[1.0pt]
 \textbf{}             & \textbf{Macro-F1}(\%) & \textbf{ACC}(\%)   & \textbf{AUC}(\%)   \\ \hline
\textbf{Wikipedia} &                       &                  &                  \\
Adaption:0         & 51.74                 & 55.03            & 51.63            \\
Adaption:1         & \textbf{93.14 }       & \textbf{91.37}   & \textbf{95.24}            \\
Adaption:2         & 92.23                 & 89.83            & 94.42            \\
Adaption:3         & 91.18                 & 88.34            & 93.33            \\
Adaption:4         & 91.90                 & 89.85            & 93.93            \\ \bottomrule[1.0pt]
\end{tabular} }
\end{center}
\label{tab:t6}
\end{table}

\subsection{Hyperparametric analysis (RQ2)}
In this subsection, we explore the effects of key hyperparameters on the performance of our model, including the number of aggregated temporal neighbours, the batch size, the embedding dimension, the span size, and the number of $N$. Our experimental results are discussed in detail below.

\textbf{Neighbor:} From Figure \ref{fig3-(a1)}, we can see that for both the DBLP and Reddit datasets, the number of neighbours does not significantly affect the prediction performance of the model. Although the experimental results show some volatility on the Wikipedida dataset, the larger number of neighbours is a more favourable choice in the overall context.

\textbf{Batchsize:} The experimental results from the three datasets in Figure \ref{fig3-(a2)} show that the prediction performance of the model increases slightly with increasing batch size in the initial stage, but after the batch size reaches 64, increasing the number of batches again decreases the performance of the model.

\textbf{Embedding:} From Figure \ref{fig3-(a3)}, we can observe that once the embedding dimension exceeds 64, there is no significant improvement in the performance of the model. Therefore, in order to minimise memory usage, it is advisable to utilize smaller embeddings whenever possible, without sacrificing the overall performance of the model.

\textbf{Span size:} Figure \ref{fig3-(a4)} clearly indicates that the model's prediction accuracy is  better  for smaller spanning steps in comparison to larger ones. This can be attributed to the fact that a smaller step size allows the model to observe the fine-grained evolution of the network, thereby resulting in improved predictive performance.

\textbf{The number of N:} In Table \ref{tab:t5}, we use the AUC criterion to evaluate the performance of DLPNN at different $N$ values (2, 4, 6, 8). From the experimental results, it can be seen that the accuracy of dynamic link prediction becomes higher as $N$ increases, but the changing values are relatively smooth. This also proves the effectiveness of our proposed model.


\subsection{Ablation studies (RQ3)}
In this subsection we conduct ablation experiments to investigate the impact of the temporal graph embedding module, the meta-learning framework, and the two types of implicit information captured on prediction performance. Overall, the predictions for these three metrics show the same trend. Below we analyse this for the AUC metric.

The first ablation experiment is to keep only the temporal graph embedding module in DLPNN, and the results are shown in Figure \ref{fig4} (w/o m-l). We can observe that the prediction performance of the temporal graph embedding module without any meta-learning adaptation is very poor on both the Wikipedia and Reddit datasets, dropping to around 50\% on wikipedia and below 90\% on Reddit.  
From the experimental results, we can conclude that the use of meta-learning techniques is effective for dynamic link prediction tasks in temporal networks.
Our second and third ablation experiments are the removal of span adaptation and node adaptation, respectively. As we can see from Figure 4(w/o en-ad) and Figure 4(w/o pre-ad), compared to the DLPNN results, node adaptation has a relatively large impact on the performance of the model, while span adaptation has a relatively small impact. This is mainly because the model has to apply the learned potential information to the new node, while span adaptation is an adaptation on the support set of the new node, so node adaptation will be more important.

\subsection{Time efficiency (RQ4)}
In this subsection, we compare the relationship between the running times of the different models in achieving optimal performance. Figure 5 shows the optimal results of four models on the three datasets and the times taken to achieve these optimal results. We can see that for both Wikipedia and Reddit, DLPNN guarantees a runtime advantage while achieving the best prediction performance. Although DyRep takes less time to reach the optimal prediction performance, its prediction performance is too poor. 
From Figure 5(c), we can see that MetaDyGNN exhibits a marginally shorter runtime compared to DLPNN. However, our prediction performance is slightly better than that of DLPNN. We attribute this phenomenon to the presence of numerous co-occurring events within the same timestamp in the DBLP dataset. As such, DLPNN requires sufficient time to acclimatise to the generation of these co-occurring events.
The results of this experiment show that our model not only has an advantage in prediction performance, but also remains comparable in terms of time consumption.

\subsection{Adaption steps (RQ4)}
In this subsection, we explore the impact of different numbers of adaption steps on the predictive performance of the model. As shown in Table \ref{tab:t6}, we run the number of adaption steps on the Wikipedia dataset from 0 to 4. The experimental results from the three evaluation criteria show that no meta-adaptation learning leads to relatively poor prediction performance. Performing one adaption step leads to the best results. When multiple adaption steps are performed, the results are still good but time consuming, so it is appropriate to choose one adaption step for DLPNN.

\section{Conclusion}
We propose DLPNN, a novel framework for dynamic link prediction of new nodes. We obtain state-of-the-art results on three datasets compared with our baselines, which demonstrate the effectiveness of DLPNN. Detailed ablation studies show that our designed meta-learning framework and the temporal graph embedding module are reasonably effective in predicting scenarios with few links in temporal graphs. This can provide a solution to scenarios with new node link prediction as well as to the node cold start problem. For future work, we will investigate how the model can be better applied to more specific scenarios, such as social platforms. In addition, the task of classifying new nodes is also worth investigating.

\section*{Acknowledgments}
The authors would like to thank all the experts who helped with this project and suggested revisions to the paper, as well as those who provided open source code and public datasets. At the same time,
 we would like to thank colleagues and experts for taking the time out of their busy schedules to provide valuable opinions.

\bibliographystyle{IEEEtran}
\bibliography{refers}

\begin{thebibliography}{10}
\providecommand{\url}[1]{#1}
\csname url@samestyle\endcsname
\providecommand{\newblock}{\relax}
\providecommand{\bibinfo}[2]{#2}
\providecommand{\BIBentrySTDinterwordspacing}{\spaceskip=0pt\relax}
\providecommand{\BIBentryALTinterwordstretchfactor}{4}
\providecommand{\BIBentryALTinterwordspacing}{\spaceskip=\fontdimen2\font plus
\BIBentryALTinterwordstretchfactor\fontdimen3\font minus \fontdimen4\font\relax}
\providecommand{\BIBforeignlanguage}[2]{{%
\expandafter\ifx\csname l@#1\endcsname\relax
\typeout{** WARNING: IEEEtran.bst: No hyphenation pattern has been}%
\typeout{** loaded for the language `#1'. Using the pattern for}%
\typeout{** the default language instead.}%
\else
\language=\csname l@#1\endcsname
\fi
#2}}
\providecommand{\BIBdecl}{\relax}
\BIBdecl

\bibitem{r49}
Y.~Zhu, F.~Lyu, C.~Hu, X.~Chen, and X.~Liu, ``Encoder-decoder architecture for supervised dynamic graph learning: A survey,'' 2022.

\bibitem{r50}
S.~M. Kazemi, R.~Goel, K.~Jain, I.~Kobyzev, A.~Sethi, P.~Forsyth, and P.~Poupart, ``Representation learning for dynamic graphs: A survey,'' 2020.

\bibitem{r44}
E.~Rossi, B.~Chamberlain, F.~Frasca, D.~Eynard, F.~Monti, and M.~Bronstein, ``Temporal graph networks for deep learning on dynamic graphs,'' \emph{arXiv preprint arXiv:2006.10637}, 2020.

\bibitem{r3}
P.~Holme and J.~Saram{\"a}ki, ``Temporal networks,'' in \emph{Encyclopedia of Social Network Analysis and Mining}, 2011.

\bibitem{r4}
Y.~Seo, M.~Defferrard, P.~Vandergheynst, and X.~Bresson, ``Structured sequence modeling with graph convolutional recurrent networks,'' in \emph{International conference on neural information processing}.\hskip 1em plus 0.5em minus 0.4em\relax Springer, 2018, pp. 362--373.

\bibitem{r8}
G.~H. Nguyen, J.~B. Lee, R.~A. Rossi, N.~K. Ahmed, E.~Koh, and S.~Kim, ``Continuous-time dynamic network embeddings,'' in \emph{Companion Proceedings of the The Web Conference 2018}, 2018, pp. 969--976.

\bibitem{r9}
Y.~Zuo, G.~Liu, H.~Lin, J.~Guo, X.~Hu, and J.~Wu, ``Embedding temporal network via neighborhood formation,'' in \emph{Proceedings of the 24th ACM SIGKDD international conference on knowledge discovery \& data mining}, 2018, pp. 2857--2866.

\bibitem{r6}
A.~Narayan and P.~H. Roe, ``Learning graph dynamics using deep neural networks,'' \emph{IFAC-PapersOnLine}, vol.~51, no.~2, pp. 433--438, 2018.

\bibitem{r7}
M.~Niepert, M.~Ahmed, and K.~Kutzkov, ``Learning convolutional neural networks for graphs,'' in \emph{International conference on machine learning}.\hskip 1em plus 0.5em minus 0.4em\relax PMLR, 2016, pp. 2014--2023.

\bibitem{r10}
D.~Xu, C.~Ruan, E.~Korpeoglu, S.~Kumar, and K.~Achan, ``Inductive representation learning on temporal graphs,'' \emph{arXiv preprint arXiv:2002.07962}, 2020.

\bibitem{r11}
E.~Rossi, B.~Chamberlain, F.~Frasca, D.~Eynard, F.~Monti, and M.~Bronstein, ``Temporal graph networks for deep learning on dynamic graphs,'' \emph{arXiv preprint arXiv:2006.10637}, 2020.

\bibitem{r41}
S.~Kumar, X.~Zhang, and J.~Leskovec, ``Predicting dynamic embedding trajectory in temporal interaction networks,'' in \emph{Proceedings of the 25th ACM SIGKDD international conference on knowledge discovery \& data mining}, 2019, pp. 1269--1278.

\bibitem{r43}
D.~Xu, C.~Ruan, E.~Korpeoglu, S.~Kumar, and K.~Achan, ``Inductive representation learning on temporal graphs,'' \emph{arXiv preprint arXiv:2002.07962}, 2020.

\bibitem{r13}
T.~Hospedales, A.~Antoniou, P.~Micaelli, and A.~Storkey, ``Meta-learning in neural networks: A survey,'' \emph{IEEE Transactions on Pattern Analysis and Machine Intelligence}, vol.~44, no.~9, pp. 5149--5169, 2022.

\bibitem{r14}
J.~Vanschoren, ``Meta-learning: A survey,'' \emph{arXiv preprint arXiv:1810.03548}, 2018.

\bibitem{r5}
M.~Defferrard, X.~Bresson, and P.~Vandergheynst, ``Convolutional neural networks on graphs with fast localized spectral filtering,'' \emph{Advances in neural information processing systems}, vol.~29, 2016.

\bibitem{r15}
F.~Zhou, C.~Cao, K.~Zhang, G.~Trajcevski, T.~Zhong, and J.~Geng, ``Meta-gnn: On few-shot node classification in graph meta-learning,'' in \emph{Proceedings of the 28th ACM International Conference on Information and Knowledge Management}, 2019, p. 2357–2360.

\bibitem{r16}
H.~Yao, C.~Zhang, Y.~Wei, M.~Jiang, S.~Wang, J.~Huang, N.~Chawla, and Z.~Li, ``Graph few-shot learning via knowledge transfer,'' in \emph{Proceedings of the AAAI Conference on Artificial Intelligence}, vol.~34, no.~04, 2020, pp. 6656--6663.

\bibitem{r17}
J.~Chauhan, D.~Nathani, and M.~Kaul, ``Few-shot learning on graphs via super-classes based on graph spectral measures,'' \emph{arXiv preprint arXiv:2002.12815}, 2020.

\bibitem{r18}
A.~J. Bose, A.~Jain, P.~Molino, and W.~L. Hamilton, ``Meta-graph: Few shot link prediction via meta learning,'' \emph{arXiv preprint arXiv:1912.09867}, 2019.

\bibitem{r29}
C.~Yang, C.~Wang, Y.~Lu, X.~Gong, C.~Shi, W.~Wang, and X.~Zhang, ``Few-shot link prediction in dynamic networks,'' in \emph{Proceedings of the Fifteenth ACM International Conference on Web Search and Data Mining}, 2022, pp. 1245--1255.

\bibitem{r48}
H.~Gerth and K.~Wolff, \emph{The Sociology of Georg Simmel}.\hskip 1em plus 0.5em minus 0.4em\relax The sociology of Georg Simmel.

\bibitem{r20}
T.~Hospedales, A.~Antoniou, P.~Micaelli, and A.~Storkey, ``Meta-learning in neural networks: A survey,'' \emph{IEEE transactions on pattern analysis and machine intelligence}, vol.~44, no.~9, pp. 5149--5169, 2021.

\bibitem{r21}
J.~X. Wang, ``Meta-learning in natural and artificial intelligence,'' \emph{Current Opinion in Behavioral Sciences}, vol.~38, pp. 90--95, 2021.

\bibitem{r22}
C.~Finn, P.~Abbeel, and S.~Levine, ``Model-agnostic meta-learning for fast adaptation of deep networks,'' in \emph{International conference on machine learning}.\hskip 1em plus 0.5em minus 0.4em\relax PMLR, 2017, pp. 1126--1135.

\bibitem{r23}
T.~N. Kipf and M.~Welling, ``Semi-supervised classification with graph convolutional networks,'' \emph{arXiv preprint arXiv:1609.02907}, 2016.

\bibitem{r19}
K.~Huang and M.~Zitnik, ``Graph meta learning via local subgraphs,'' \emph{Advances in neural information processing systems}, vol.~33, pp. 5862--5874, 2020.

\bibitem{r30}
A.~Pareja, G.~Domeniconi, J.~Chen, T.~Ma, T.~Suzumura, H.~Kanezashi, T.~Kaler, T.~Schardl, and C.~Leiserson, ``Evolvegcn: Evolving graph convolutional networks for dynamic graphs,'' in \emph{Proceedings of the AAAI Conference on Artificial Intelligence}, vol.~34, no.~04, 2020, pp. 5363--5370.

\bibitem{r31}
J.~Chen, X.~Wang, and X.~Xu, ``Gc-lstm: Graph convolution embedded lstm for dynamic link prediction,'' \emph{arXiv preprint arXiv:1812.04206}, 2018.

\bibitem{r33}
W.~Jin, H.~Jiang, M.~Qu, T.~Chen, C.~Zhang, P.~A. Szekely, and X.~Ren, ``Recurrent event network : Global structure inference over temporal knowledge graph,'' \emph{arXiv: Learning}, 2019.

\bibitem{r34}
Y.~Seo, M.~Defferrard, P.~Vandergheynst, and X.~Bresson, ``Structured sequence modeling with graph convolutional recurrent networks,'' in \emph{International conference on neural information processing}.\hskip 1em plus 0.5em minus 0.4em\relax Springer, 2018, pp. 362--373.

\bibitem{r38}
Y.~Zuo, G.~Liu, H.~Lin, J.~Guo, X.~Hu, and J.~Wu, ``Embedding temporal network via neighborhood formation,'' in \emph{Proceedings of the 24th ACM SIGKDD international conference on knowledge discovery \& data mining}, 2018, pp. 2857--2866.

\bibitem{r39}
G.~H. Nguyen, J.~B. Lee, R.~A. Rossi, N.~K. Ahmed, E.~Koh, and S.~Kim, ``Continuous-time dynamic network embeddings,'' in \emph{Companion proceedings of the the web conference 2018}, 2018, pp. 969--976.

\bibitem{r40}
Y.~Wang, Y.-Y. Chang, Y.~Liu, J.~Leskovec, and P.~Li, ``Inductive representation learning in temporal networks via causal anonymous walks,'' \emph{arXiv preprint arXiv:2101.05974}, 2021.

\bibitem{r42}
R.~Trivedi, M.~Farajtabar, P.~Biswal, and H.~Zha, ``Dyrep: Learning representations over dynamic graphs,'' in \emph{International conference on learning representations}, 2019.

\bibitem{r47}
Y.~Lu, X.~Wang, C.~Shi, P.~S. Yu, and Y.~Ye, ``Temporal network embedding with micro-and macro-dynamics,'' in \emph{Proceedings of the 28th ACM international conference on information and knowledge management}, 2019, pp. 469--478.

\bibitem{r45}
W.~Hamilton, Z.~Ying, and J.~Leskovec, ``Inductive representation learning on large graphs,'' \emph{Advances in neural information processing systems}, vol.~30, 2017.

\bibitem{r46}
P.~Velickovic, G.~Cucurull, A.~Casanova, A.~Romero, P.~Lio, Y.~Bengio \emph{et~al.}, ``Graph attention networks,'' \emph{stat}, vol. 1050, no.~20, pp. 10--48\,550, 2017.

\end{thebibliography}

\vfill

\end{document}